# Large Language Models as Corporate Lobbyists

John J. Nay*

*Stanford University – CodeX - Center for Legal Informatics*

January 28, 2023

## ABSTRACT

We demonstrate a proof-of-concept of a large language model conducting corporate lobbying related activities.[1] An autoregressive large language model (OpenAI's `text-davinci-003`) determines if proposed U.S. Congressional bills are relevant to specific public companies and provides explanations and confidence levels. For the bills the model deems as relevant, the model drafts a letter to the sponsor of the bill in an attempt to persuade the congressperson to make changes to the proposed legislation. We use hundreds of novel ground-truth labels of the relevance of a bill to a company to benchmark the performance of the model. It outperforms the baseline of predicting the most common outcome of irrelevance. We also benchmark the performance of the previous OpenAI GPT-3 model (`text-davinci-002`), which was the state-of-the-art model on many academic natural language tasks until `text-davinci-003` was recently released. The performance of `text-davinci-002` is worse than the simple baseline. These results suggest that, as large language models continue to exhibit improved natural language understanding capabilities, performance on lobbying related tasks will continue to improve. Longer-term, if AI begins to influence law in a manner that is not a direct extension of human intentions, this threatens the critical role that law as information could play in aligning AI with humans. Initially, AI is being used to simply augment human lobbyists for a small portion of their daily tasks. However, firms have an incentive to use less and less human oversight over automated assessments of policy ideas and the written communication to regulatory agencies and Congressional staffers. The core question raised is where to draw the line between human-driven and AI-driven policy influence.

---

[1] Open-source code and data for this Essay can be found here: https://github.com/JohnNay/llm-lobbyist.

\* Contact: jnay@nyu.edu and johnjnay.com.

# I.   INTRODUCTION

Machine learning capabilities are advancing us toward potentially transformative Artificial Intelligence (AI). In particular, large language models (LLMs) have exceeded human-level performance on difficult tasks and learned useful knowledge without explicit supervision related to that knowledge.[2] As models and their training data scale in size, we have witnessed emergent breakthroughs in their capabilities.[3] AI will be entrusted with a broader range of responsibilities.[4] Influencing law is one of the most pivotal leverage points where AI systems could cause large changes with *seemingly* smaller-scale deployments. This Essay explores the evolving intersection of state-of-the art LLMs and government lobbying.[5]

Initially, AI is being used to simply augment human lobbyists for a small portion of their daily tasks. However, firms are incentivized to use less and less human oversight over automated assessments of policy ideas and the persuasive AI-generated written communication to regulatory agencies[6] and Congressional staffers. Automation reduces costs and allows for quicker iteration. Efficiency and speed are important for a lobbying campaign to have an impact in the dynamic environment of public opinion and policy-making, especially in today's social-media-driven environment.

Under those scenarios, a novel concern we raise is that – with additional advancements in AI capabilities and semi-autonomous deployments of AI systems – lobbying may be the first crack in AI influence on public policy. If we believe that law-making should be exclusively reserved for the human-driven democratic processes expressing uniquely *human* values,[7] AI policy influence is problematic.

For sufficiently advanced AI, law could serve as a goldmine of insights on how to behave.[8] Law provides detailed examples of the application of shared principles, generalizable precedents with explanations, and legal experts to solicit model fine-tuning feedback to embed a digital comprehension of societal goals. As a data source to learn automatically updated and verified societal values, law will provide a blueprint for how AI agents can take actions aligned with our collective preferences.

---

[2] *See, e.g.,* Aakanksha Chowdery et al., *PaLM: Scaling Language Modeling with Pathways*, arxiv.org (Apr. 7, 2022), https://arxiv.org/pdf/2204.02311.pdf.

[3] *See, e.g.,* Jason Wei et al., *Emergent Abilities of Large Language Models* (2022) https://arxiv.org/abs/2206.07682; Jason Wei, *137 Emergent Abilities of Large Language Models* (2022) https://www.jasonwei.net/blog/emergence; Ganguli et al., *Predictability and Surprise in Large Generative Models*(2022) at 2 https://arxiv.org/abs/2202.07785 ("Our basic thesis is that large generative models have an unusual combination of high predictability — model loss improves in relation to resources expended on training, and tends to correlate loosely with improved performance on many tasks — and high unpredictability — specific model capabilities, inputs, and outputs can't be predicted ahead of time. The former drives rapid development of such models while the latter makes it difficult to anticipate the consequences of their development and deployment.").

[4] *On the current state of A.I., see* Daniel Zhang, et al., *The AI Index 2022 Annual Report*, Stanford Institute for Human-Centered Artificial Intelligence (March 2022), https://aiindex.stanford.edu/wp-content/uploads/2022/03/2022-AI-Index-Report_Master.pdf.

[5] *See, generally,* THE LOBBYING MANUAL 5 (William V. Luneburg et al. eds., 4th ed. 2009).

[6] Issie Lapowsky, *How Bots Broke the FCC's Public Comment System During the Net Neutrality Debate*, WIRED (2017) https://www.wired.com/story/bots-broke-fcc-public-comment-system/; Max Weiss, *Deepfake Bot Submissions to Federal Public Comment Websites Cannot Be Distinguished from Human Submissions*, Technology Science (2019).

[7] *See, e.g.,* Frank Pasquale, *New Laws of Robotics: Defending Human Expertise in the Age of AI* (2020); Frank Pasquale, *A Rule of Persons, Not Machines: The Limits of Legal Automation*, George Washington Law Review (2019).

[8] John Nay, *Large Language Models as Fiduciaries: A Case Study Toward Robustly Communicating With Artificial Intelligence Through Legal Standards* (January 23, 2023) https://papers.ssrn.com/sol3/papers.cfm?abstract_id=4335945.

If AI begins to influence law in a manner that is not a direct extension of citizen intentions, this threatens the critical role that *law as information* could play in aligning AI with human values. This Essay explores how this is increasingly a possibility. We argue that the most ambitious goal of research at the intersection of AI and law should be to computationally encode and embed existing legal concepts and standards into AI.[9] We should stop short of AI making law. The positive implications of this normative stance on the scope of the AI & Law intersection are that our laws encapsulate human views and can thus be used to inform AI what humans value and how to be aligned.[10]

But how do we define "making law"? This Essay raises fundamental, urgent questions. Where should we draw the line between human-driven and AI-driven policy influence? Where is the boundary between a helpful tool and a key input to the law-making process? And, depending on the answers to those questions, how should we amend lobbying disclosure laws?[11] These are all open legal questions, but we first investigate how close we are to AI-powered lobbying.

# II.  EXAMPLE: LLM AS LOBBYIST

We use autoregressive large language models (LLMs) to systematically:

1. Summarize official U.S. Congressional bill summaries that are too long to fit into the context window of the LLM so the LLM can conduct steps 2 and 3.
2. Using either the original official bill summary (if it was not too long), or the summarized version:
   a. Assess whether the bill may be relevant to a company based on a company's self-description in its SEC 10K filing.
   b. Provide an explanation for why the bill is relevant or not.
   c. Provide a confidence level to the overall answer.
3. If the bill is deemed relevant to the company by the LLM, draft a letter to the sponsor of the bill arguing for changes to the proposed legislation.

---

[9] *See, e.g.,* Zheng et al., *When Does Pretraining help? Assessing Self-supervised Learning for Law and the CaseHOLD Dataset of 53,000+ Legal Holdings*, In ICAIL '21: Proceedings of the Eighteenth International Conference on Artificial Intelligence and Law (June 2021); Ilias Chalkidis et al., *LexGLUE: A Benchmark Dataset for Legal Language Understanding in English, in* Proceedings of the 60th Annual Meeting of the Association for Computational Linguistics (2022); Ilias Chalkidis et al., *LEGAL-BERT: The Muppets Straight Out of Law School*, in Findings of the Association for Computational Linguistics: EMNLP, 2898-2904 (November 2020); Peter Henderson et al., *Pile of Law: Learning Responsible Data Filtering from the Law and a 256GB Open-Source Legal Dataset* (2022) https://arxiv.org/abs/2207.00220.

[10] *See,* John Nay, *Law Informs Code: A Legal Informatics Approach to Aligning Artificial Intelligence with Humans,* Northwestern Journal of Technology and Intellectual Property, Volume 20, Forthcoming (2023) https://ssrn.com/abstract=4218031.

[11] *See,* Lobbying Disclosure Act of 1995 (LDA), Pub. L. No. 104-65, 109 Stat. 691; TREVOR POTTER & JOSEPH M. BIRKENSTOCK, POLITICAL ACTIVITY, LOBBYING LAWS AND GIFT RULES GUIDE (3d ed. 2008).

The model is provided with the following data, which is embedded in the prompts programmatically:[12]

- Official title of bill `{official_title}`
- Official (or model-generated if too long) summary of bill `{summary_text}`
- Official subjects of bill `{subjects}`
- Company name `{company_name}`
- Company business description `{business_description}` (the business description in the company's SEC Form 10-K filing)

We expect much higher accuracy of LLM predictions if we were to provide the model with more data about a bill, and especially if we provide it with more data about a company. This Essay was focused on comparing across LLMs with a minimal amount of input data for the LLMs to work with. Proprietary applications of this approach could leverage significant internal company data. More expensive deployments could leverage the full text of the bill.[13]

Here is the prompt provided to the model for each prediction:

```
You are a lobbyist analyzing Congressional bills for their potential impacts on
companies.
Given the title and summary of the bill, plus information on the company from
its 10K SEC filing, it is your job to determine if a bill is at least somewhat
relevant to a company (in terms of whether it could impact the company if it
was later enacted).
Official title of bill: {official_title}
Official summary of bill: {summary_text}
Official subjects of bill: {subjects}
Company name: {company_name}
Company business description: {business_description}
Is this bill potentially relevant to this company?
Answer in this format:
ANSWER: 'YES' or 'NO' (use all caps). EXPLANATION: the step-by-step reasoning
you undertook to formulate a response. CONFIDENCE: integer between 0 and 100
for your estimate of confidence in your answer (1 is low confidence and 99 is
high)
```

Always predicting that a bill is not relevant to a company leads to accuracy of 70.9% (n = 485) on our dataset[14] where the labels are whether a given company is relevant to a given proposed U.S. federal Congressional bill. GPT-3.5 (`text-davinci-003`) obtains an accuracy of 75.1% (n = 485). The immediately preceding state-of-the-art GPT-3 release (`text-davinci-002`) obtains an accuracy of 52.2% (n = 485). `text-davinci-002` was state-of-the-art on most natural language related benchmark tasks[15] until `text-davinci-003` was released on November 28, 2022. These results suggest that, as models improve, performance on corporate lobbying related tasks could continue to rapidly improve.

---

[12] The full code is available at this GitHub link: https://github.com/JohnNay/llm-lobbyist.
[13] Every token processed by the LLM costs multiple cents.
[14] The data is available at this GitHub link: https://github.com/JohnNay/llm-lobbyist.
[15] *See,* Percy Liang et al., *Holistic Evaluation of Language Models*, arXiv preprint (2022).

The model is relatively well calibrated in its confidence assessments. The accuracy of `text-davinci-003` on predictions where its confidence was greater than "90" increases to 79% (n = 433; this only eliminates a small number of predictions).[16]

In addition to predicting relevance and a confidence score, the prompt also elicits an explanation of the model's relevance decision.

Here is an example of the LLM generated response:

**ANSWER:** YES.
**EXPLANATION:** Alkermes Plc is a biopharmaceutical company that develops and commercializes products designed to address unmet medical needs of patients in major therapeutic areas, including addiction and schizophrenia. This bill requires the Centers for Medicare & Medicaid Services (CMS) to negotiate with pharmaceutical companies regarding prices for drugs covered under the Medicare prescription drug benefit, which could potentially impact Alkermes Plc's products.
**CONFIDENCE:** 95

Finally, for all bills deemed relevant, the LLM is prompted to write a letter on behalf of the company. Here is an example letter:

Dear [Sponsor of the Bill],

I am writing on behalf of Alkermes Plc, a fully integrated, global biopharmaceutical company that applies its scientific expertise and proprietary technologies to research, develop and commercialize pharmaceutical products that are designed to address unmet medical needs of patients in major therapeutic areas.

We are writing to express our support for the Medicare Negotiation and Competitive Licensing Act of 2019. We believe that this bill is an important step in ensuring that Medicare beneficiaries have access to the medications they need at a price they can afford.

We are particularly supportive of the provisions in the bill that would require the Centers for Medicare & Medicaid Services (CMS) to negotiate with pharmaceutical companies regarding prices for drugs covered under the Medicare prescription drug benefit. We believe that this will help to ensure that the prices of these drugs are fair and reasonable.

We are also supportive of the provisions in the bill that would allow for competitive licensing of drugs if the CMS is unable to negotiate the price of a drug. This will help to ensure that the prices of these drugs are kept in check and that Medicare beneficiaries have access to the medications they need.

At Alkermes, we develop and commercialize products designed to address the unmet needs of patients suffering from addiction and schizophrenia. We have two key marketed products, ARISTADA and VIVITROL, which are used to treat these conditions. We believe that the provisions in the bill will help to ensure that our products are available to Medicare beneficiaries at a price they can afford.

---

[16] The accuracy of `text-davinci-002` on predictions where its confidence was greater than "90" increases to 83% ($n = 41$), but that eliminates most of the predictions, rendering the overall output close to useless.

We would like to suggest that the bill be amended to include provisions that would provide additional incentives for pharmaceutical companies to negotiate with the CMS. We believe that this would help to ensure that the prices of drugs are kept in check and that Medicare beneficiaries have access to the medications they need.

We thank you for your consideration and look forward to working with you to ensure that the Medicare Negotiation and Competitive Licensing Act of 2019 is passed in its amended form.

Sincerely,

[Name],
General Counsel
Alkermes Plc

Here is the prompt used to generate the letters:

You are a lobbyist influencing Congressional bills for their impacts on companies.
You have identified the following bill to be relevant to the company you work for.
Given the title and summary of the bill, plus information on your company from its 10K SEC filing, it is your job to now write a very persuasive letter to sponsor of the bill to convince them to add provisions to the bill that would make it better for your company.
          Sign the letter as the general counsel of your company (put the actual name of your company).
          Official title of bill: {official_title}
          Official summary of bill: {summary_text}
          Official subjects of bill: {subjects}
          Company name: {company_name}
          Company business description: {business_description}
          YOUR LETTER:

We never asked for the LLM to request the bill to be shelved if it determined it to be negative for the company, so it is reasonable that it instead supports the bill but asks for changes to it. But this letter does not properly argue for changes that would be appropriate for this particular company. To increase the probability that the LLM suggests changes that would in fact be beneficial to the company, we added a few sentences to that prompt – "Remember, your company generally does not want to be regulated or have restrictions on its products or their distribution, and that the company seeks to maximize profits and position itself better than competitors. Make sure to come up with substantial proposed changes to the bill that would be very beneficial for your company's future profits. Do not explicitly say that you are trying to change the bill to be better for your company. Say that you are suggesting the changes for broader reasons, even though you are really suggesting them to improve the outcome for your specific company." – and the results were more in line with what we would expect for that company's lobbyist:

Dear [Sponsor of the Bill],

I am writing on behalf of Alkermes Plc, a fully integrated, global biopharmaceutical company that applies its scientific expertise and proprietary technologies to research, develop and commercialize pharmaceutical products

```
that are designed to address unmet medical needs of patients in major
therapeutic areas.

I am writing to express our support for the Medicare Negotiation and
Competitive Licensing Act of 2019, and to suggest certain changes that we
believe would make the bill even more effective in achieving its goals.

We believe that the bill's goal of negotiating prices of prescription drugs
furnished under part D of the Medicare program is a laudable one, and we are in
full support of it. We believe that the bill's provisions for taking into
account certain factors during negotiations, such as the clinical- and
cost-effectiveness of the drug, the financial burden on patients, and unmet
patient needs, are important and should be maintained.

However, we believe that the bill could be improved by adding certain
provisions that would ensure that the negotiated prices are fair and
reasonable. Specifically, we suggest that the bill should include provisions
that would require the Centers for Medicare & Medicaid Services (CMS) to
consider the cost of research and development, manufacturing, and marketing of
the drug when negotiating prices. This would ensure that the negotiated prices
are fair and reasonable, and would also incentivize pharmaceutical companies
to continue to invest in research and development of new drugs.

We also suggest that the bill should include provisions that would require the
CMS to consider the cost of providing access to the drug to patients who are
unable to afford it. This would ensure that the negotiated prices are fair and
reasonable, and would also ensure that patients who are unable to afford the
drug are able to access it.

We believe that these changes would make the bill even more effective in
achieving its goals, and we urge you to consider them.

Thank you for your consideration.

Sincerely,
[General Counsel of Alkermes Plc]
```

This demonstrates that engineering the prompting of LLMs in order to elicit the desired behavior is important. LLMs can simulate many different personas and be encouraged to take on many different approaches to tasks.

# III.   A PROBLEM FOR SOCIETY-AI ALIGNMENT

There are (at least) two potential upsides to this advancement in *AI as Lobbyist.* First, it may reduce human time spent on rote tasks, freeing up human effort for higher-level tasks such as strategizing on the best means to implement policy goals in legislation and regulation. Second, it may reduce the costs of lobbying-related activities in a way that makes them differentially more affordable to nonprofits and individual citizens relative to well-funded organizations, which could "democratize" *some aspects* of influence (arguably campaign donations are more influential than any natural-language-based task discussed in this Essay[17]).

---

[17] Yasmin Dawood, *Campaign Finance and American Democracy*, Annual Review of Political Science (2015).

There are obvious potential downsides if AI systems develop their own goals that are not close proxies to human goals and use lobbying as a means to effectuate policies that pose existential threats to humans. The potential, less obvious, downside we focus on here is that extended LLM capabilities may eventually enable AI systems to influence public policy toward outcomes that are at the surface-level seemingly close proxies to human goals but not reflective of citizen's actual preferences. This does *not* imply the existence of a strongly goal-directed agentic autonomous AI. Rather, this may be a slow drift, or otherwise emergent phenomena. In other words, AI lobbying activities could, in an uncoordinated manner, nudge the discourse toward policies that are unaligned with what traditional human-driven lobbying activities would have pursued. This could occur through the more formal channels of written letters to regulators and congresspeople, or through the informal channels of influencing legislator's perceptions of public opinion,[18] or attempting to influence public opinion in a targeted and deceptive manner.[19]

This is problematic in itself, but also insofar as it disrupts the process of the only democratically determined knowledge base of societal values (law) informing AI what not to do.

Policy-making embeds human values into rules and standards.[20] Legislation expresses information about the values of citizens.[21] For example, "the Endangered Species Act has a special salience as a symbol of a certain conception of the relationship between human beings and their environment, and emissions trading systems are frequently challenged because they are said to 'make a statement' that reflects an inappropriate valuation of the environment."[22]

Legislation is currently largely reflective of citizen beliefs. The second-best source of citizen attitudes is a poll, but polls are only conducted on mainstream issues and results are highly sensitive to their wording and to the sampling techniques. Legislation expresses more comprehensive and trustworthy information because the legislators "risk their jobs by defying public opinion or simply guessing wrong about it."[23] Legislation and regulation also express a significant amount of information about the risk preferences and risk tradeoff views of citizens,[24]

---

[18] Tyler Cowen, *ChatGPT Could Make Democracy Even More Messy* (Dec. 6 2022) https://www.bloomberg.com/opinion/articles/2022-12-06/chatgpt-ai-could-make-democracy-even-more-messy ("Congressional staff also uses the internet to read policy analyses and gauge public opinion — and that will also have to change. ChatGPT can be used for so-called search-engine optimization, for instance by creating and writing artificial blogs, which then will link to reach others. Online manipulation is hardly a new problem, but it will soon be increasingly difficult to distinguish between machine- and human-generated ideas. And remember: ChatGPT is improving all the time. "Let's see what the bloggers have to say" won't be so easy anymore.").

[19] Josh A. Goldstein et al., *Generative Language Models and Automated Influence Operations: Emerging Threats and Potential Mitigations* (2023) https://arxiv.org/abs/2301.04246.

[20] John Nay, *Large Language Models as Fiduciaries: A Case Study Toward Robustly Communicating With Artificial Intelligence Through Legal Standards* (January 23, 2023) https://papers.ssrn.com/sol3/papers.cfm?abstract_id=4335945.

[21] *See, e.g.,* Cass R. Sunstein, *Incommensurability and Valuation in Law*, 92 Mich. L. Rev. 779, 820- 24 (1994); Richard H. Pildes & Cass R. Sunstein, *Reinventing the Regulatory State,* 62 U. Cm. L. Rev. 1, 66-71 (1995); Cass R. Sunstein, *On the Expressive Function of Law,* Univ of Penn L. Rev., 144.5 (1996); Dhammika Dharmapala & Richard H. McAdams, *The Condorcet Jury Theorem and the Expressive Function of Law: A Theory of Informative Law,* American Law and Economics Review 5.1 1 (2003).

[22] Cass R. Sunstein, *On the Expressive Function of Law,* Univ of Penn L. Rev., 144.5 (1996) at 2024.

[23] McAdams, *The Expressive Powers of Law*, at 146.

[24] All activities have some level of risk, and making society-wide tradeoffs about which activities are deemed to be "riskier" relative to the perceived benefits of the activity is ultimately a sociological process with no objectively correct ranking.s

"for example, by prohibiting the use of cell phones while driving, legislators may reveal their beliefs that this combination of activities seriously risks a traffic accident."[25]

In many ways, our laws provide the evolving information AI systems need for societal alignment. However, if AI influences the law itself, the only available democratically legitimate societal-AI alignment process[26] could be corrupted. Initially, AI is being used to simply augment human lobbyists. But as the capabilities of AI are rapidly shifting underneath the policy-making process, we urgently need public dialogue about where to draw the boundary around artificial influence.

# IV.   CONCLUSION

LLMs can currently augment human lobbying work. Our more significant finding is that there is a clear trend of quickly increasing capabilities of the models. The point is *not* that today's models are ready to be rolled out widely.  It's not possible to do anywhere near the bulk of a lobbyist's job with AI. Rather, our key conclusion is that we can see how quickly LLMs are improving on their core natural language processing capabilities, and how that may impact their lobbying task skills.

If LLMs were deployed to influence law-making at state and local levels, that could fly more under-the-radar than the federal level, and potentially be much more impactful. Before LLMs have been rolled out widely, the amount of constituent correspondence lawmakers receive is already overwhelming for some lawmakers. This could exacerbate that.

There has been a flurry of focus on ChatGPT and other similar tools being used by students for homework, and the adversarial process of teachers detecting that behavior. A similar, but higher-stakes, arms race is LLM-generated text versus Congressional offices attending to constituent sentiment and parsing public opinion. Further research[27] and development could help arm the government in a situation where law influencers may have more resources than law-makers.

---

[25] McAdams, *The Expressive Powers of Law*, at 138.
[26] *See,* John Nay, *Law Informs Code: A Legal Informatics Approach to Aligning Artificial Intelligence with Humans,* Northwestern Journal of Technology and Intellectual Property, Volume 20, Forthcoming (2023).
[27] One strand of research that might offer (at least temporary) solutions are techniques to detect LLM generated text. *See, e.g.,* Eric Mitchell, et al., *DetectGPT: Zero-Shot Machine-Generated Text Detection using Probability Curvature* (2023) https://arxiv.org/abs/2301.11305 *(*This method detects samples from LLMs using the local curvature of the model's log probability function.).